\documentclass[runningheads]{llncs}
\usepackage[T1]{fontenc}
\usepackage{graphicx}
\usepackage{cite}
\usepackage{amsmath,amssymb,amsfonts}
\usepackage{esvect}
\usepackage{algorithmic}
\usepackage{textcomp}
\usepackage{xcolor}
\usepackage{booktabs}
\usepackage{multirow}
\usepackage{pifont}
\usepackage{diagbox}
\usepackage[caption=false]{subfig}

\begin{document}
\title{\LARGE{TS-P$^2$CL: Plug-and-Play Dual Contrastive Learning for Vision-Guided Medical Time Series Classification}}
%
\titlerunning{TS-P$^2$CL: Dual Contrastive Learning for Brain Signal Analysis}
%
\author{Qi'ao Xu\inst{1}\textsuperscript{†}\orcidID{0009-0002-5567-438X} \and Pengfei Wang\inst{1}\textsuperscript{†} \and Bo Zhong\inst{1} \and Tianwen Qian\inst{1}\textsuperscript{\ding{41}}\orcidID{0000-0002-3881-4857} \and Xiaoling Wang\inst{1}\textsuperscript{\ding{41}} \and Ye Wang\inst{2} \and Hong Yu\inst{2}
}
\authorrunning{Q. Xu et al.}
%
\institute{East China Normal University, Shanghai, 200062, China \\
\email{\{qaxu,pfwang\}@stu.ecnu.edu.cn; \{twqian,xlwang\}@cs.ecnu.edu.cn}\\
\and
Chongqing University of Posts and Telecommunications, Chongqing, 400065, China \\
}
%
\maketitle              

\begin{abstract}
Medical time series (MedTS) classification is pivotal for intelligent healthcare, yet its efficacy is severely limited by poor cross-subject generation due to the profound cross-individual heterogeneity. Despite advances in architectural innovations and transfer learning techniques, current methods remain constrained by modality-specific inductive biases that limit their ability to learn universally invariant representations. To overcome this, we propose TS-P$^2$CL, a novel plug-and-play framework that leverages the universal pattern recognition capabilities of pre-trained vision models. We introduce a vision-guided paradigm that transforms 1D physiological signals into 2D pseudo-images, establishing a bridge to the visual domain. This transformation enables implicit access to rich semantic priors learned from natural images. Within this unified space, we employ a dual-contrastive learning strategy: intra-modal consistency enforces temporal coherence, while cross-modal alignment aligns time-series dynamics with visual semantics, thereby mitigating individual-specific biases and learning robust, domain-invariant features. Extensive experiments on six MedTS datasets demonstrate that TS-P$^2$CL consistently outperforms fourteen methods in both subject-dependent and subject-independent settings.


\end{abstract}

\section{Introduction}
\label{label:introduction}

Medical time series (MedTS), such as electroencephalography (EEG), electrocardiography (ECG), and stereo-electroencephalography (sEEG), capture the dynamic changes of physiological signals over time \cite{qian2025deep}. The automated classification of these signals holds immense potential for intelligent healthcare, enabling real-time monitoring, early intervention, and personalized medicine.
However, a fundamental challenge impedes their clinical deployment: poor cross-subject generalization. Models trained on one cohort often degrade significantly when applied to new individuals, due to substantial cross-subject variability in signal morphology, amplitude, and noise profiles. This heterogeneity renders representations learned from limited, modality-specific data prone to overfitting, undermining robustness in real-world settings.

The research community has explored two primary directions: 1) architectural innovations using deep neural networks (CNNs, RNNs, and Transformers~\cite{wang2024medformer}) to model complex temporal dependencies, and 2) transfer learning techniques involving either large-scale time series pretraining or language model adaptation~\cite{liang2024foundation}.
Despite notable progress, these approaches face inherent limitations. The former relies heavily on modality-specific inductive biases, which can hinder the learning of universally invariant representations. The latter, harnessing powerful pre-trained language knowledge, suffers from a {fundamental modality gap} when projecting continuous signals into discrete text, potentially distorting temporal dynamics.   

Inspired by the remarkable universal pattern recognition capabilities of pre-trained vision models, we explore a vision-guided paradigm for MedTS classification by treating physiological signals as pseudo-images. Our key insight is that time series and natural images share deep, intrinsic structural commonalities: trends manifest as edges, periodicities resemble textures, and abrupt changes correspond to discontinuities. This suggests that the powerful, general-purpose representations learned by vision models like ViT\cite{dosovitskiy2020image} and VanillaNet~\cite{chen2024vanillanet} can be transferred to understand physiological dynamics.
This thought is not only theoretically grounded but also clinically intuitive: physicians routinely diagnose by visually inspecting waveforms~\cite{li2024visual}. Recent studies have demonstrated the efficacy of this paradigm in time series forecasting~\cite{chen2024visionts, zhong2025time}, but they are primarily designed for forecasting and lack explicit mechanisms for learning robust representations for classification under significant cross-subject heterogeneity.

In this paper, we propose \textbf{TS-P$^2$CL}, a plug-and-play MedTS classification method that improves cross-subject generalization through dual contrastive learning. 
We transform 1D physiological signals into 2D pseudo-images, bridging them into the rich feature space of pre-trained vision models. 
To learn robust and generalizable representations, we employ two complementary objectives: (\textit{i}) \textit{Intra-modal consistency} ensures temporal coherence by contrasting different augmentations of the same time series. (\textit{ii}) \textit{Cross-modal alignment} directly aligns the time-series embeddings with the frozen features of a pre-trained vision model, facilitating the acquisition of semantically meaningful visual representations.
Crucially, our approach requires no fine-tuning of the vision model, enabling efficient, plug-and-play deployment while leveraging off-the-shelf visual knowledge. 

Our contributions are summarized as follows:
\begin{itemize}
    \item We introduce {TS-P$^2$CL}, a plug-and-play framework that bridges the gap between vision-based zero-shot transfer and MedTS representation learning, enhancing generalization through dual contrastive learning.
    \item We develop a vision-guided paradigm where a frozen pre-trained vision model serves as a semantic anchor for cross-modal alignment, enabling efficient knowledge transfer without any modification.
    \item We conduct extensive experiments on six MedTS datasets, showing that TS-P$^2$CL outperforms fourteen methods under both subject- and subject-independent setups, validating its effectiveness and strong generalization.
\end{itemize}


\section{Related Work}
\label{label:related_work}
\subsection{Medical Time Series Classification}
Medical time series (MedTS) classification plays a crucial role in disease diagnosis and health monitoring through physiological signal analysis. These signals are often noisy, high-dimensional, and exhibit significant cross-subject variability, posing challenges for robust classification. Classical methods include dynamic time warping and shapelets \cite{liang2021efficient}, which are interpretable but limited in scalability. Statistical models like autoregressive models \cite{kaur2023autoregressive} capture temporal dynamics but struggle with non-stationarity. Recently, deep learning has driven significant progress. For example, Medformer~\cite{wang2024medformer} employs a multi-granularity transformer to model complex temporal patterns. These advances reflect progress toward more robust and clinically relevant MedTS models.

\subsection{Cross-Modal Learning}
Cross-modal learning leverages complementary signals from different modalities to extract rich semantic features. Existing studies have shown that pre-training across modalities can produce highly transferable representations for downstream tasks~\cite{kim2025comprehensive}. Notable examples include the audio-visual contrastive learning and CLIP’s joint image-text embedding space~\cite{radford2021learning}, which learn unified representations by contrasting matched and mismatched pairs. This paradigm has also gained traction in healthcare, especially in brain signal analysis, where visual or auditory cues can enhance feature learning. For instance, aligning EEG with fMRI through self-supervised improves representative robustness~\cite{wei2025multi}. These efforts highlight the potential of cross-modal methods to bridge heterogeneous data and improve generalization in real-world applications.

\section{Method}
\label{label:method}

\begin{figure*}[!ht]
\centering
\includegraphics[width=0.90\linewidth]{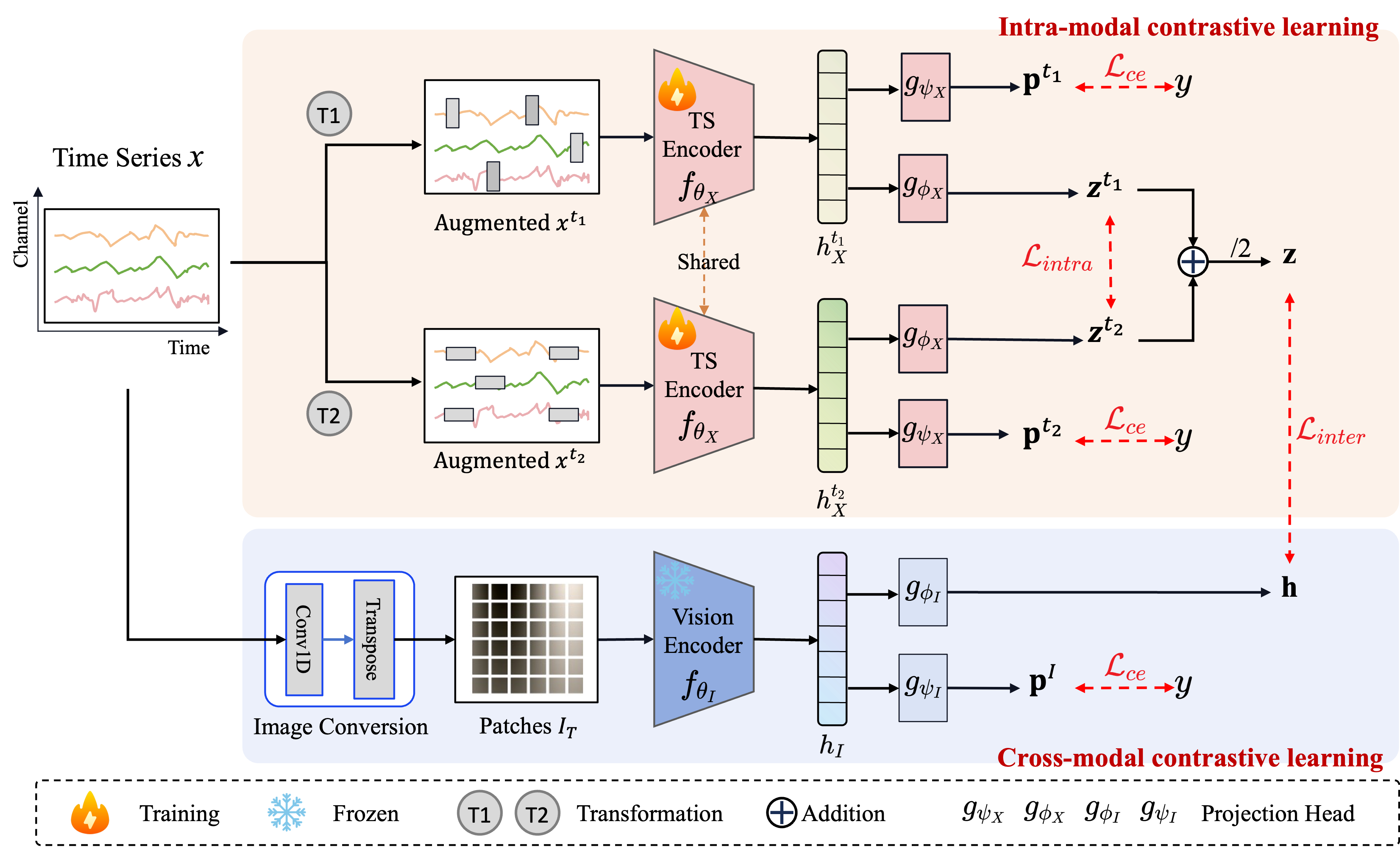}
\caption{Overall architecture of TS-P$^2$CL. The framework employs dual contrastive learning during training: (\textit{a}) {Intra-modal contrastive learning} ensures temporal robustness by contrasting augmented views of the time series, and (\textit{b}) {Cross-modal contrastive learning} aligns time-series embeddings with visual features from a frozen pre-trained vision model, transferring universal visual knowledge.
During inference, only the time series branch is applied. 
}
\label{fig:framework}
\vspace{-1em}
\end{figure*}

\subsection{Overall Framework}
\label{label:overall}
Given a medical time series dataset $D = \{(x_i, y_i)\}_{i=1}^{N}$, where  $x_i \in \mathbb{R}^{C \times T} $ denotes a multivariate time series and $y_i$ its corresponding label, the goal of MedTS classification is to learn a robust encoder $f_{\theta_{X}}$ and a projection head $g_{\phi_X}$ enables accurate prediction.  
We propose {TS-P$^2$CL}, a plug-and-play framework that enhances cross-subject generalization through dual contrastive learning with visual guidance. As shown in Fig.~\ref{fig:framework}, our approach consists of three key components:

\begin{itemize}
\item {Intra-modal contrastive learning} ($\mathcal{L}_{\text{intra}}$): By treating differently augmented views of the same input as positive pairs, it enforces temporal consistency and reduce semantic drift, yielding stable and discriminative representations.
\item {Cross-modal contrastive learning with visual guidance} ($\mathcal{L}_{\text{inter}}$): Transforms 1D signals into 2D pseudo-images and align their prototypes with visual features from a {frozen} pre-trained vision model (\textit{e.g.}, ViT, EfficientNet), transferring universal visual knowledge to enhance generalization.
\item {Progressive joint learning}: Enables end-to-end optimization by jointly learning representation and classification objectives, with a schedule that shifts focus from contrastive learning to task-specific prediction.
\end{itemize}

\subsection{Intra-Modal Contrastive Learning}
To enhance the stability and discriminative capability of time series representations, we employ intra-modal contrastive learning. This approach constructs positive-negative pairs via multi-view augmentation, aligning augmented views from the same input while separating them from others in feature space. 

\subsubsection{Multi-View Positive Pair Construction}
For each input $x_i$, we generate two augmented views $(x_i^{t_1},x_i^{t_2})$ by randomly sampling two spatial-temporal masks $t_1, t_2$ from a transformation set $T$ \cite{zhong2023ts}. 
Each view is encoded and projected by: 
\begin{align}
    h_{X,i}^t = f_{\theta_{X}}(x_i^t), 
    \quad    
    \mathbf{z}_i^{t} = g_{\phi_X}(h_{X,i}^t), 
    \quad
    \text{where } t\in\{t_1,t_2\}
\label{eq:ts_embedding}
\end{align}
where $f_{\theta_X}(\cdot)$ denotes a TS encoder and $g_{\phi_X}(\cdot)$ is a projection head.

\subsubsection{Intra-Modal Contrastive Loss}
Within a mini-batch $\mathcal{B} = \{x_i\}_{i=1}^{B}$, we treat $\{\mathbf{z}_i^{t_1}, \mathbf{z}_i^{t_2}\}$ as a positive pair and all remaining embeddings $\{\mathbf{z}_j^{t_1}, \mathbf{z}_j^{t_2}\}_{j \neq i}$ as negatives. Using the NT-Xent loss~\cite{chen2020simple}, we maximize the similarity of positive pair relative to all negatives:
\begin{equation}
l\left(i, t_1, t_2\right)=-\log \frac{\exp \left(s\left(\mathbf{z}_i^{t_1}, \mathbf{z}_i^{t_2}\right) / \tau\right)}
{\sum_{\substack{k=1 \\ k \neq i}}^B \exp \left(s\left(\mathbf{z}_i^{t_1}, \mathbf{z}_k^{t_1}\right) / \tau\right)+\sum_{k=1}^B \exp \left(s\left(\mathbf{z}_i^{t_1}, \mathbf{z}_k^{t_2}\right) / \tau\right)},
\label{eq:intra_cl_loss}
\end{equation}
where $s(\cdot, \cdot)$ denotes cosine similarity and temperature $\tau > 0$ jointly calibrate the separation margin, with smaller $\tau$ sharpening the positive-negative boundary.

The intra-modal contrastive loss is computed as follows:
\begin{equation}
    \mathcal{L}_{intra}=\frac{1}{2 N} \sum_{i=1}^N\left[l\left(i, t_1, t_2\right)+l\left(i, t_2, t_1\right)\right].
\label{eq:intra_loss}
\end{equation}
It minimizes intra-series discrepancy under varying transforms while maximizing inter-series separation, thereby sharpening the model’s discriminability.

\subsection{Cross-Modal Contrastive Learning with Visual Guidance}
To enhance the generalization capability of time series representations, we employ cross-modal contrastive learning that aligns time series features with visual embeddings from pre-trained vision models. 
\vspace{-1em}

\subsubsection{Time-Series Prototype Generation}
For reliable cross-modal alignment, we construct a unified and view-invariant time-series prototype $\mathbf{z}_i$ by averaging the augmented embeddings $\{\mathbf{z}_i^{t_1}, \mathbf{z}_i^{t_2}\}$ from the intra-modal stage:
 \begin{equation}
 \vspace{-0.25em}
    \mathbf{z}_i =  \left( \mathbf{z}_i^{t_1} + \mathbf{z}_i^{t_2} \right) / 2.
\vspace{-0.25em}
\end{equation} 
This prototype integrates consistent temporal patterns across views and serves as a stable anchor for alignment with visual representations.

\vspace{-1em}

\subsubsection{Visual Representation Extraction}
To bridge the modality gap, we convert the 1D time series into 2D pseudo-images using a learnable image-conversion module tailored to different vision architectures.
\textit{For CNN-based encoders} (\textit{e.g.}, EfficientNet \cite{tan2019efficientnet}), we reshape the input $X \in \mathbb{R}^{C \times T}$ into a single-channel 2D tensor $X' \in \mathbb{R}^{1 \times C \times T}$, and apply 2D convolution to generate pseudo-images $I_C$:
\begin{equation}
\vspace{-0.25em}
    I_C = \text{Conv2D}(X') \in \mathbb{R}^{C' \times H' \times W'},
\vspace{-0.25em}
\end{equation}
where $C'$, $H'$, $W'$ are the output channels, height, and width, respectively. 

\textit{For Transformer-based models} (\textit{e.g.}, ViT~\cite{dosovitskiy2020image}), we generate $N = \lfloor (T - P)/S \rfloor + 1$ overlapping patches $I_T$ via 1D convolution with kernel size $P$, stride $S$, and output dimension $D$:
\begin{equation}
\vspace{-0.25em} 
    I_T = \text{Transpose}(\text{Conv1D}(X)) \in \mathbb{R}^{N \times D},
\vspace{-0.25em}
\end{equation}
which are treated as visual tokens by the ViT encoder.

The generated pseudo-image $I \in \{I_C, I_T\}$ is then processed by a frozen pre-trained vision encoder $f_{\theta_I}(\cdot)$ and projected into the contrastive space via $g_{\phi_I}(\cdot)$:
\begin{equation}
    h_I = f_{\theta_I}(I) \in \mathbb{R}^{D}, \quad \mathbf{h}_i = g_{\phi_I}(h_I) \in \mathbb{R}^{D'},
\end{equation}
where $\mathbf{h}_i$ is the final visual feature. Crucially, the vision encoder $f_{\theta_I}(\cdot)$ is kept {frozen}, ensuring our framework is plug-and-play and computationally efficient.

\vspace{-1em}

\subsubsection{Cross-Modal Contrastive Loss}
We treat $(\mathbf{z}_i, \mathbf{h}_i)$ as a positive pair, assuming they represent the same underlying physiological state in different modalities. All other embeddings $\{\mathbf{z}_j, \mathbf{h}_k\}_{j,k \neq i}$ form negatives.
The cross-modal contrastive loss is defined using the NT-Xent loss:
\begin{equation}
c(i, \mathbf{z}, \mathbf{h})=-\log \frac{\exp \left(s\left(\mathbf{z}_i, \mathbf{h}_i\right) / \tau\right)}
{\sum_{\substack{k=1 \\ k \neq i}}^B \exp \left(s\left(\mathbf{z}_i, \mathbf{z}_k\right) / \tau\right) + \sum_{k=1}^B \exp \left(s\left(\mathbf{z}_i, \mathbf{h}_k\right) / \tau\right)}
\end{equation}

The cross-modal contrastive loss is computed as follows:
\begin{equation}
    \mathcal{L}_{inter} = \frac{1}{2 N} \sum_{i=1}^N \left[ c(i, \mathbf{z}, \mathbf{h}) + c(i, \mathbf{h}, \mathbf{z}) \right].
\label{eq:inter_loss}
\end{equation}
This aligns time-series and visual prototypes while suppressing domain-specific noise, thereby amplifying cross-modal discriminability.

\subsection{Progressive Joint Learning}
To enable end-to-end optimization and harmonize representation learning with task-specific discrimination, we introduce a progressive joint learning strategy. Unlike conventional two-stage paradigms, our approach dynamically balances self-supervised contrastive objectives and supervised classification, allowing the model to first discover robust structures and then refine predictive accuracy.

\subsubsection{Dual Contrastive Objective}
We define the dual contrastive loss $\mathcal{L}_{\text{dual}}$ as the sum of intra- and cross-modal components:
\begin{equation}
    \mathcal{L}_{dual} = \mathcal{L}_{intra} + \mathcal{L}_{inter},
\end{equation}
where $\mathcal{L}_{intra}$ enforces temporal consistency across augmented views (Eq.~\ref{eq:intra_loss}), and $\mathcal{L}_{inter}$ aligns time-series prototypes with frozen visual features (Eq.~\ref{eq:inter_loss}). 
This objective guides the time series encoder $f_{\theta_X}$ to learn representations that are both temporally consistent and semantically aligned with universal visual knowledge.

\subsubsection{Task-Specific Classification}
The classification objective employs a standard cross-entropy loss $\mathcal{L}_{ce}$ to enhance discriminative learning across both modalities:
\begin{equation}
    \mathbf{p}^I = g_{\psi_I}(h_I), 
    \quad
    \mathbf{p}^{t} = g_{\psi_X}(h_X^{t}),
    \quad 
    \text{where } t \in\{t_1, t_2\}
\end{equation}
\begin{equation}
    \mathcal{L}_{cls} = \frac{1}{3N} \sum_{i=1}^{N} (\mathcal{L}_{ce}(\mathbf{p}^I_i, y_i) + \mathcal{L}_{ce}( \mathbf{p}^{t_1}_i, y_i) + \mathcal{L}_{ce}(\mathbf{p}^{t_2}_i, y_i )),
\end{equation}
where $g_{\psi_v}$ and $g_{\psi_s}$ are modality-specific classifiers, and $y_{i}$ is the ground truth.

\begin{table}[!ht]
\centering
\caption{Summary of six physiological time series datasets, including the number of subjects, sample, channels, sampling rate, classes, and modality.}
\label{tab:physiological_datasets}
\resizebox{0.7\linewidth}{!}{
\begin{tabular}{c|cccccc}
\toprule
{Dataset} & {Subject} & {Samples} & {Channel} & {Sampling Rate} & {Class} & {Modality} \\ 
\midrule
ADFD & 88 & 5,967 & 19 & 256Hz & 3 & EEG \\ 
APAVA & 23 & 69,752 & 16 & 256Hz & 2 & EEG\\ 
TDBRAIN & 72 & 6,240 & 33 & 256Hz & 3 & EEG\\ 
PTB & 198 & 64,356 & 15 & 250Hz & 2 & ECG\\ 
PTB-XL & 17,596 & 191,400 & 12 & 250Hz & 5 & ECG\\ 
sEEG & 5 & 471,991 &102-146 & 1024Hz & 2 & sEEG \\
\bottomrule
\end{tabular}
}
\vspace{-1em} 
\end{table}

\subsubsection{Adaptive Learning Schedule}
The total training objective combines representation and classification learning via adaptive weighting:
\begin{equation}
\vspace{-0.25em}
    \mathcal{L} = \lambda \cdot \mathcal{L}_{dual} + (1-\lambda) \cdot \mathcal{L}_{cls},
    \quad
    \text{where } \lambda(T) = 1 - \left( \frac{T}{T_{\text{Max}}} \right)^2,
    \label{eq:total_loss}
\vspace{-0.25em}
\end{equation}
where $T$ is the current epoch and $T_{Max}$ is the total number of epochs. This strategy prioritizes contrastive learning in early stages to build robust representations, gradually shifting focus to classification for fine-grained decision refinement.

\section{Experiment}
\label{label:experiment}

\subsection{Experimental Setups}
\label{label:experimental_setups}

\subsubsection{Datasets}
\label{label:datasets}
We conduct experiments on six clinical time series datasets across EEG, ECG, and sEEG modalities. The EEG group includes ADFD~\cite{miltiadous2023dataset}, APAVA~\cite{escudero2006analysis}, and TDBRAIN~\cite{van2022two}. The ECG group contains PTB~\cite{physiobank2000physionet} and PTB-XL~\cite{wagner2020ptb}. The sEEG dataset is a private collection from five patients at Huashan Hospital~\cite{yu2021characterizing}, with recordings labeled as pre-ictal abnormal or normal sleep-rest. More details are provided in Table~\ref{tab:physiological_datasets}.
All datasets are split into train/validation/test sets with a 6:2:2 ratio. Evaluation follows two protocols: \ding{172} \textit{Subject-dependent}: Samples are split randomly across subjects. \ding{173} \textit{Subject-independent}: All trials from a subject remain in one split. Unless specified, we adopt the subject-independent protocol.

\vspace{-1em}

\subsubsection{Baselines}
\label{label:baselines}
We compare with four categories of time series methods:
1) {Transformer-based}: FEDformer~\cite{zhou2022fedformer}, PatchTST~\cite{nie2022pathtst}, ViT~\cite{dosovitskiy2020image}, Medformer~\cite{wang2024medformer}, and iTransformer~\cite{liu2023itransformer}.
2) {MLP-based}: DLinear~\cite{zeng2023dlinear} and TSMixer~\cite{chen2023tsmixer}.
3) {CNN-based}: EfficientNet~\cite{tan2019efficientnet} and VanillaNet~\cite{chen2024vanillanet}.
4) {Semi-supervised}: MatchingNets~\cite{vinyals2016matching}, RelationNets~\cite{sung2018learning}, PrototypicalNets~\cite{snell2017prototypical}, MAML~\cite{finn2017model}, and PiSC~\cite{yu2021characterizing}.

\vspace{-1em}

\begin{table*}[!ht]
\centering
\caption{Subject-dependent classification performance on ADFD dataset. Best results are marked in \textbf{bold}, and second-best are \underline{underlined}. All subsequent tables follow this convention.}
\scalebox{0.85}{
\begin{tabular}{l | c | c | c | c | c }
        \toprule
        {Models} & Accuracy & {Precision} & {Recall} & {F1 score} & {AUROC} \\
        \midrule
        FEDformer & 78.79$\scriptstyle\pm0.90$ & 77.76$\scriptstyle\pm0.84$ & 77.98$\scriptstyle\pm0.88$ & 77.81$\scriptstyle\pm0.86$ & 92.37$\scriptstyle\pm0.50$ \\
        DLinear & 39.18$\scriptstyle\pm0.57$ & 34.95$\scriptstyle\pm0.65$ & 34.39$\scriptstyle\pm0.46$ & 32.69$\scriptstyle\pm0.59$ & 50.35$\scriptstyle\pm0.29$ \\
        PatchTST & 66.38$\scriptstyle\pm0.53$ & 65.30$\scriptstyle\pm0.55$ & 65.28$\scriptstyle\pm0.70$ & 65.14$\scriptstyle\pm0.53$ & 83.19$\scriptstyle\pm0.29$ \\
        iTransformer & 65.12$\scriptstyle\pm0.27$ & 62.74$\scriptstyle\pm0.23$ & 62.51$\scriptstyle\pm0.33$ & 62.50$\scriptstyle\pm0.35$ & 81.61$\scriptstyle\pm0.21$ \\
        TSMixer & 93.81$\scriptstyle\pm0.58$ & 93.50$\scriptstyle\pm0.62$ & 93.50$\scriptstyle\pm0.63$ & 93.49$\scriptstyle\pm0.60$ & 99.02$\scriptstyle\pm0.16$ \\
        Medformer & \underline{97.61$\scriptstyle\pm0.18$} & \underline{97.65$\scriptstyle\pm0.12$} & \underline{97.35$\scriptstyle\pm0.25$} & \underline{97.50$\scriptstyle\pm0.17$} & \underline{99.83$\scriptstyle\pm0.03$} \\
        TS-P$^2$CL (\textbf{Ours}) & \textbf{98.36$\scriptstyle\pm0.05$} & \textbf{98.32$\scriptstyle\pm0.03$} & \textbf{98.31$\scriptstyle\pm0.06$} & \textbf{98.29$\scriptstyle\pm0.11$} & \textbf{99.91$\scriptstyle\pm0.12$} \\
        \bottomrule
    \end{tabular}
    }
\label{tab:sub-dep-ADFD}
\vspace{-1em}
\end{table*}

\begin{table}[!ht]
\centering
\caption{Subject-dependent classification performance on sEEG dataset.}
\label{tab:seeg_subject_dependent}
\scalebox{0.82}{
\begin{tabular}{l|c|c}
    \toprule
    Models &  Accuracy & F1 score \\
    \midrule
    MatchingNets & 51.23$\scriptstyle\pm1.68$ & 52.12$\scriptstyle\pm2.96$ \\
    RelationNets & 68.32$\scriptstyle\pm0.71$ & 71.70$\scriptstyle\pm1.00$ \\
    PrototypicalNets & 69.96$\scriptstyle\pm1.92$ & 71.44$\scriptstyle\pm1.63$ \\
    MAML & 75.58$\scriptstyle\pm0.79$ & 76.87$\scriptstyle\pm1.29$ \\
    PiSC & 84.99$\scriptstyle\pm0.97$ & 85.26$\scriptstyle\pm0.83$ \\
    \midrule
    EfficientNet-B0 & 85.59$\scriptstyle\pm1.60$ & 85.65$\scriptstyle\pm1.64$ \\
    EfficientNet-B0 + $\mathcal{L}_{{dual}}$ & 89.75$\scriptstyle\pm1.20$ & 89.95$\scriptstyle\pm1.34$ \\
    \midrule
    ViT-Tiny & 86.62$\scriptstyle\pm0.65$ & 86.75$\scriptstyle\pm1.67$ \\
    ViT-Tiny + $\mathcal{L}_{{dual}}$ & \underline{90.97$\scriptstyle\pm0.85$} & \underline{91.05$\scriptstyle\pm0.86$} \\
    \midrule
    VanillaNet & 87.31$\scriptstyle\pm0.70$ & 87.35$\scriptstyle\pm1.31$ \\
    VanillaNet + $\mathcal{L}_{{intra}}$ & 88.53$\scriptstyle\pm1.25$ & 88.65$\scriptstyle\pm0.77$ \\
    VanillaNet + $\mathcal{L}_{{inter}}$ & 90.20$\scriptstyle\pm0.80$ & 90.30$\scriptstyle\pm1.02$ \\
    VanillaNet + $\mathcal{L}_{{dual}}$ (\textbf{Ours}) & \textbf{91.74$\scriptstyle\pm0.90$} & \textbf{91.75$\scriptstyle\pm0.91$} \\
    \bottomrule
\end{tabular}}
\vspace{-1em}
\end{table}

\subsubsection{Implementation Details}
\label{label:implementation_details}
We adopt five standard evaluation metrics: accuracy, precision, recall, F1 score, and AUROC. 
All experiments are conducted on an NVIDIA RTX A6000 GPU (48GB) with PyTorch 2.4 and CUDA 11.8. 
We employ Adam optimizer with a learning rate of $5\times10^{-4}$ and dataset-specific batch sizes $({32, 32, 128, 128, 128, 256})$. Training stopped early if validation F1 plateaued for 10 epochs, and results are averaged over five random seeds.
For deployment efficiency, all vision-branch components \( f_{\theta_I}(\cdot) \), \( g_{\phi_I}(\cdot) \), \( g_{\psi_I}(\cdot) \), and \( g_{\psi_X}(\cdot) \) are removed after training. Inference relies solely on the TS encoder \( f_{\theta_X}(\cdot) \) and projection head \( g_{\phi_X}(\cdot) \), ensuring a compact and deployable architecture.

\subsection{Main Results}
\label{label:main_results}

\subsubsection{Subject-Dependent Analysis}
Table~\ref{tab:sub-dep-ADFD} and Table~\ref{tab:seeg_subject_dependent} show the subject-dependent classification performance on ADFD and sEEG datasets. 
\ding{172} On ADFD, TS-P$^2$CL outperforms all baselines, achieving \(98.36\%\) accuracy, \(98.32\%\) precision, \(98.31\%\) recall, \(98.29\%\) F1, and \(99.91\%\) AUROC. 
\ding{173} On sEEG, TS-P$^2$CL achieves \(91.74\%\) accuracy and \(91.75\%\) F1, surpassing PiSC and vision-based methods, demonstrating stronger generalization. 
These findings indicate that our method achieves superior discrimination under consistent data distributions.

\vspace{-1em}

\begin{table*}[!ht]
\centering
\caption{Subject-independent classification performance on five public datasets.}
\label{tab:multi_dataset_results}
\scalebox{0.80}{
\begin{tabular}{c|c|c|c|c|c|c|c|c}
\toprule
{Dataset} & {Metric} & \textbf{TS-P$^2$CL} & Medformer & TSMixer & iTransformer & PatchTST & DLinear & FEDformer \\
\midrule
    \multirow{5}{*}{\rotatebox{90}{ADFD}}
    & Accuracy  & \textbf{56.08$\scriptstyle\pm0.14$} & \underline{53.19$\scriptstyle\pm1.25$} & 51.61$\scriptstyle\pm1.07$ & 52.07$\scriptstyle\pm1.36$ & 43.85$\scriptstyle\pm0.52$ & 40.15$\scriptstyle\pm0.61$ & 46.25$\scriptstyle\pm0.78$ \\
    & Precision & \textbf{55.13$\scriptstyle\pm0.12$} & \underline{52.10$\scriptstyle\pm1.85$} & 49.23$\scriptstyle\pm1.88$ & 46.83$\scriptstyle\pm0.96$ & 42.30$\scriptstyle\pm1.42$ & 36.28$\scriptstyle\pm0.62$ & 45.14$\scriptstyle\pm1.81$ \\
    & Recall    & \textbf{55.66$\scriptstyle\pm0.62$} & \underline{50.19$\scriptstyle\pm1.02$} & 48.23$\scriptstyle\pm1.80$ & 47.02$\scriptstyle\pm1.31$ & 40.77$\scriptstyle\pm1.51$ & 34.28$\scriptstyle\pm0.23$ & 43.98$\scriptstyle\pm1.46$ \\
    & F1 score  & \textbf{55.15$\scriptstyle\pm0.55$} & \underline{50.51$\scriptstyle\pm1.02$} & 48.36$\scriptstyle\pm1.93$ & 46.42$\scriptstyle\pm1.02$ & 40.01$\scriptstyle\pm1.98$ & 32.79$\scriptstyle\pm0.50$ & 44.02$\scriptstyle\pm1.52$ \\
    & AUROC     & \textbf{75.05$\scriptstyle\pm0.73$} & \underline{70.78$\scriptstyle\pm1.56$} & 68.22$\scriptstyle\pm1.07$ & 66.85$\scriptstyle\pm1.06$ & 60.01$\scriptstyle\pm0.94$ & 50.19$\scriptstyle\pm0.20$ & 61.91$\scriptstyle\pm1.66$ \\
\midrule
    \multirow{5}{*}{\rotatebox{90}{APAVA}}
    & Accuracy  & \textbf{83.05$\scriptstyle\pm0.56$} & 74.12$\scriptstyle\pm8.14$ & \underline{76.67$\scriptstyle\pm5.86$} & 75.12$\scriptstyle\pm1.35$ & 68.50$\scriptstyle\pm1.95$ & 53.66$\scriptstyle\pm0.41$ & 75.07$\scriptstyle\pm2.41$ \\
    & Precision & \textbf{87.75$\scriptstyle\pm0.18$} & 71.28$\scriptstyle\pm6.15$ & \underline{77.71$\scriptstyle\pm7.29$} & 74.85$\scriptstyle\pm1.53$ & 78.05$\scriptstyle\pm2.03$ & 50.31$\scriptstyle\pm0.43$ & 74.61$\scriptstyle\pm1.57$ \\
    & Recall    & \textbf{79.05$\scriptstyle\pm0.12$} & 72.04$\scriptstyle\pm7.12$ & \underline{73.76$\scriptstyle\pm5.94$} & 72.88$\scriptstyle\pm2.04$ & 62.19$\scriptstyle\pm2.70$ & 50.27$\scriptstyle\pm0.38$ & 74.11$\scriptstyle\pm4.52$ \\
    & F1 score  & \textbf{80.51$\scriptstyle\pm0.56$} & 72.18$\scriptstyle\pm7.06$ & \underline{74.48$\scriptstyle\pm6.29$} & 73.78$\scriptstyle\pm2.15$ & 59.51$\scriptstyle\pm3.91$ & 49.79$\scriptstyle\pm0.40$ & 73.25$\scriptstyle\pm3.65$ \\
    & AUROC     & \underline{86.01$\scriptstyle\pm0.51$} & 77.52$\scriptstyle\pm5.12$ & \textbf{86.03$\scriptstyle\pm7.15$} & 85.07$\scriptstyle\pm2.38$ & 66.56$\scriptstyle\pm3.49$ & 49.57$\scriptstyle\pm0.42$ & 84.05$\scriptstyle\pm2.02$ \\
\midrule
    \multirow{5}{*}{\rotatebox{90}{PTB}}
    & Accuracy  & \textbf{85.02$\scriptstyle\pm1.14$} & 80.14$\scriptstyle\pm1.88$ & \underline{82.09$\scriptstyle\pm1.05$} & 84.15$\scriptstyle\pm2.46$ & 75.98$\scriptstyle\pm1.68$ & 74.41$\scriptstyle\pm1.43$ & 77.11$\scriptstyle\pm1.95$ \\
    & Precision & \underline{86.97$\scriptstyle\pm0.21$} & 82.31$\scriptstyle\pm1.02$ & \textbf{85.77$\scriptstyle\pm2.37$} & 87.69$\scriptstyle\pm1.72$ & 79.47$\scriptstyle\pm2.61$ & 74.89$\scriptstyle\pm1.99$ & 79.04$\scriptstyle\pm3.91$ \\
    & Recall    & \textbf{78.05$\scriptstyle\pm0.68$} & 72.10$\scriptstyle\pm2.89$ & \underline{74.30$\scriptstyle\pm1.73$} & 75.73$\scriptstyle\pm3.55$ & 65.93$\scriptstyle\pm1.88$ & 64.09$\scriptstyle\pm2.26$ & 66.72$\scriptstyle\pm3.28$ \\
    & F1 score  & \textbf{81.20$\scriptstyle\pm1.91$} & 75.42$\scriptstyle\pm2.61$ & \underline{76.62$\scriptstyle\pm1.78$} & 79.02$\scriptstyle\pm4.11$ & 66.95$\scriptstyle\pm2.01$ & 64.74$\scriptstyle\pm2.83$ & 69.89$\scriptstyle\pm4.13$ \\
    & AUROC     & \textbf{93.35$\scriptstyle\pm0.65$} & \underline{92.93$\scriptstyle\pm1.28$} & 91.78$\scriptstyle\pm1.03$ & 91.01$\scriptstyle\pm2.86$ & 88.21$\scriptstyle\pm1.48$ & 88.21$\scriptstyle\pm0.55$ & 86.70$\scriptstyle\pm3.62$ \\
\midrule
    \multirow{5}{*}{\rotatebox{90}{PTB-XL}}
    & Accuracy  & \textbf{74.01$\scriptstyle\pm0.51$} & \underline{72.61$\scriptstyle\pm0.84$} & 70.42$\scriptstyle\pm0.30$ & 68.91$\scriptstyle\pm0.58$ & 73.51$\scriptstyle\pm0.75$ & 45.49$\scriptstyle\pm0.02$ & 59.71$\scriptstyle\pm9.19$ \\
    & Precision & \textbf{66.53$\scriptstyle\pm0.70$} & \underline{64.21$\scriptstyle\pm0.63$} & 60.87$\scriptstyle\pm0.44$ & 59.88$\scriptstyle\pm0.59$ & 65.97$\scriptstyle\pm0.59$ & 21.25$\scriptstyle\pm8.71$ & 55.32$\scriptstyle\pm8.11$ \\
    & Recall    & \textbf{61.64$\scriptstyle\pm0.81$} & \underline{60.33$\scriptstyle\pm0.65$} & 55.67$\scriptstyle\pm0.12$ & 54.21$\scriptstyle\pm0.41$ & 60.21$\scriptstyle\pm0.61$ & 20.05$\scriptstyle\pm0.04$ & 49.85$\scriptstyle\pm7.79$ \\
    & F1 score  & \textbf{62.72$\scriptstyle\pm0.24$} & \underline{61.93$\scriptstyle\pm0.88$} & 57.52$\scriptstyle\pm0.03$ & 56.48$\scriptstyle\pm0.40$ & 62.28$\scriptstyle\pm0.41$ & 12.63$\scriptstyle\pm0.12$ & 49.58$\scriptstyle\pm8.69$ \\
    & AUROC     & \textbf{90.55$\scriptstyle\pm0.61$} & \underline{89.81$\scriptstyle\pm0.63$} & 87.66$\scriptstyle\pm0.18$ & 86.79$\scriptstyle\pm0.31$ & 89.39$\scriptstyle\pm0.54$ & 50.62$\scriptstyle\pm0.09$ & 84.13$\scriptstyle\pm4.79$ \\
\midrule
    \multirow{5}{*}{\rotatebox{90}{TDBRAIN}}
    & Accuracy  & \textbf{92.01$\scriptstyle\pm1.05$} & \underline{88.16$\scriptstyle\pm1.19$} & 79.52$\scriptstyle\pm1.33$ & 76.13$\scriptstyle\pm5.09$ & 74.07$\scriptstyle\pm4.49$ & 55.90$\scriptstyle\pm1.01$ & 77.11$\scriptstyle\pm2.05$ \\
    & Precision & \underline{88.15$\scriptstyle\pm0.75$} & \textbf{88.91$\scriptstyle\pm1.63$} & 79.81$\scriptstyle\pm1.28$ & 75.08$\scriptstyle\pm1.35$ & 76.08$\scriptstyle\pm5.17$ & 56.68$\scriptstyle\pm1.18$ & 78.80$\scriptstyle\pm2.41$ \\
    & Recall    & \textbf{91.21$\scriptstyle\pm0.80$} & \underline{89.11$\scriptstyle\pm1.69$} & 79.52$\scriptstyle\pm1.33$ & 75.03$\scriptstyle\pm1.81$ & 74.05$\scriptstyle\pm5.41$ & 55.90$\scriptstyle\pm1.01$ & 77.69$\scriptstyle\pm1.24$ \\
    & F1 score  & \textbf{91.41$\scriptstyle\pm1.08$} & \underline{87.79$\scriptstyle\pm2.15$} & 79.47$\scriptstyle\pm1.34$ & 75.50$\scriptstyle\pm1.94$ & 74.07$\scriptstyle\pm4.67$ & 54.59$\scriptstyle\pm1.43$ & 77.84$\scriptstyle\pm1.59$ \\
    & AUROC     & \underline{96.34$\scriptstyle\pm0.06$} & \textbf{96.79$\scriptstyle\pm0.72$} & 88.85$\scriptstyle\pm1.36$ & 84.61$\scriptstyle\pm1.48$ & 81.30$\scriptstyle\pm7.21$ & 59.41$\scriptstyle\pm1.10$ & 87.15$\scriptstyle\pm1.55$ \\
\bottomrule
\end{tabular}
}
\label{tab:sub-indep-all}
\vspace{-1.5em}
\end{table*}

\begin{table}[!ht]
\centering
\begin{minipage}[t]{0.46\textwidth}
\centering
\caption{Subject-independent classification accuracy on sEEG dataset.}
\scalebox{0.90}{
\begin{tabular}{l|c}
\toprule
Model Variants & Accuracy \\
\midrule
VanillaNet  & 54.86 \\
VanillaNet + $\mathcal{L}_{{inter}}$ & 63.37 \\
VanillaNet + $\mathcal{L}_{{intra}}$  & 64.58 \\
\midrule
Efficientnet-B0 + $\mathcal{L}_{{dual}}$  & 63.74 \\
ViT-Tiny + $\mathcal{L}_{{dual}}$   & 65.46 \\
VanillaNet + $\mathcal{L}_{{dual}}$ (\textbf{Ours}) & 69.23 \\
\bottomrule
\end{tabular}}
\label{tab:seeg_domain}
\end{minipage}
\hfill
\begin{minipage}[t]{0.50\textwidth}
\centering
\caption{Pairwise cross-subject generalization accuracy on sEEG dataset.}
\scalebox{0.85}{
\begin{tabular}{c|ccccc|c}
\toprule
\diagbox{Src}{Tgt} & P1 & P2 & P3 & P4 & P5 & \textbf{Avg.} \\
\midrule
P1 & - & 84.81 & 35.64 & 91.05 & 91.28 & 75.70 \\
P2 & 80.14 & - & 64.53 & 90.44 & 50.39 & 71.38 \\
P3 & 77.76 & 81.80 & - & 91.05 & 91.22 & 85.46 \\
P4 & 29.22 & 49.96 & 37.25 & - & 26.45 & 35.72 \\
P5 & 78.30 & 83.47 & 76.55 & 73.30 & - & 77.90 \\
\midrule
\textbf{Avg.} & 66.35 & 75.01 & 53.49 & 86.46 & 69.65 & 69.23 \\
\bottomrule
\end{tabular}}
\label{table:seeg_domain_detail}
\end{minipage}
\vspace{-2em}
\end{table}

\subsubsection{Subject-Independent Analysis}
Table~\ref{tab:sub-indep-all} presents the subject-independent performance on five public datasets. TS-P$^2$CL achieves the best results on 21 of 25 metrics and second-best on remaining four. 
This consistent F1 dominance reflects a strong precision-recall balance, indicating robust cross-subject generalization. Such balance is clinically critical: minimizing false positives (misdiagnosing healthy subjects) avoids unnecessary treatments and psychological distress, while reducing false negatives (missing pathological cases) helps prevent delayed interventions. 
These results show TS-P$^2$CL offers superior error calibration and generalization under subject shift, making it suitable for practical deployment.

\vspace{-1em}

\begin{figure}
    \centering
    \includegraphics[width=0.90\linewidth]{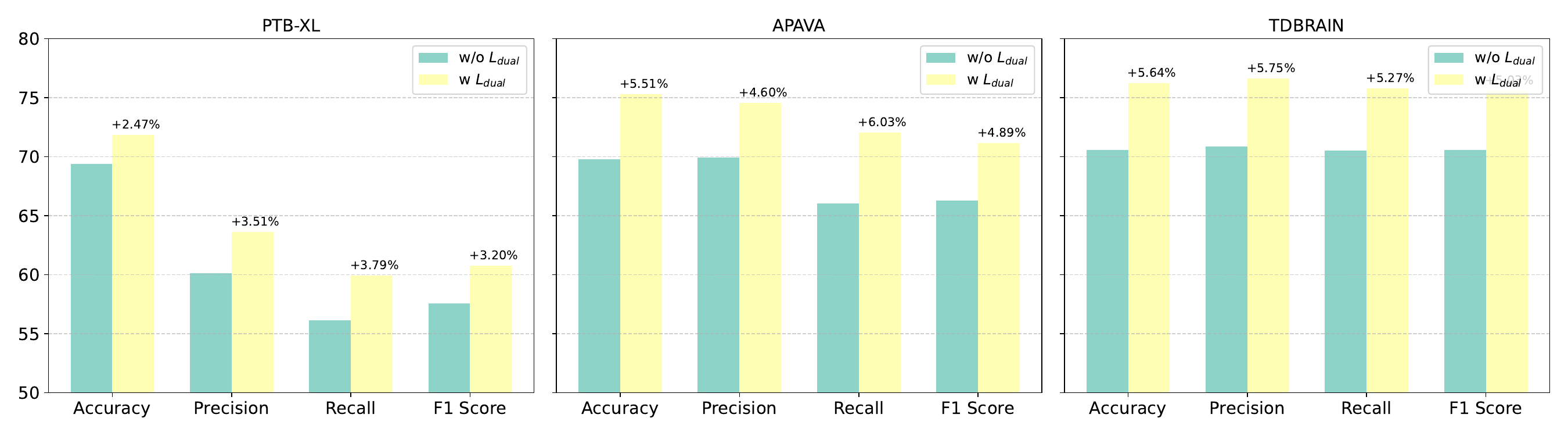}
    \caption{Ablation study of $\mathcal{L}_{dual}$ on PTB-XL, APAVA, and TDBRAIN datasets.}
    \label{fig:abl_dual_loss}
\vspace{-1em}
\end{figure}


\begin{figure}[!ht]
    \centering
    \includegraphics[width=0.78\linewidth]{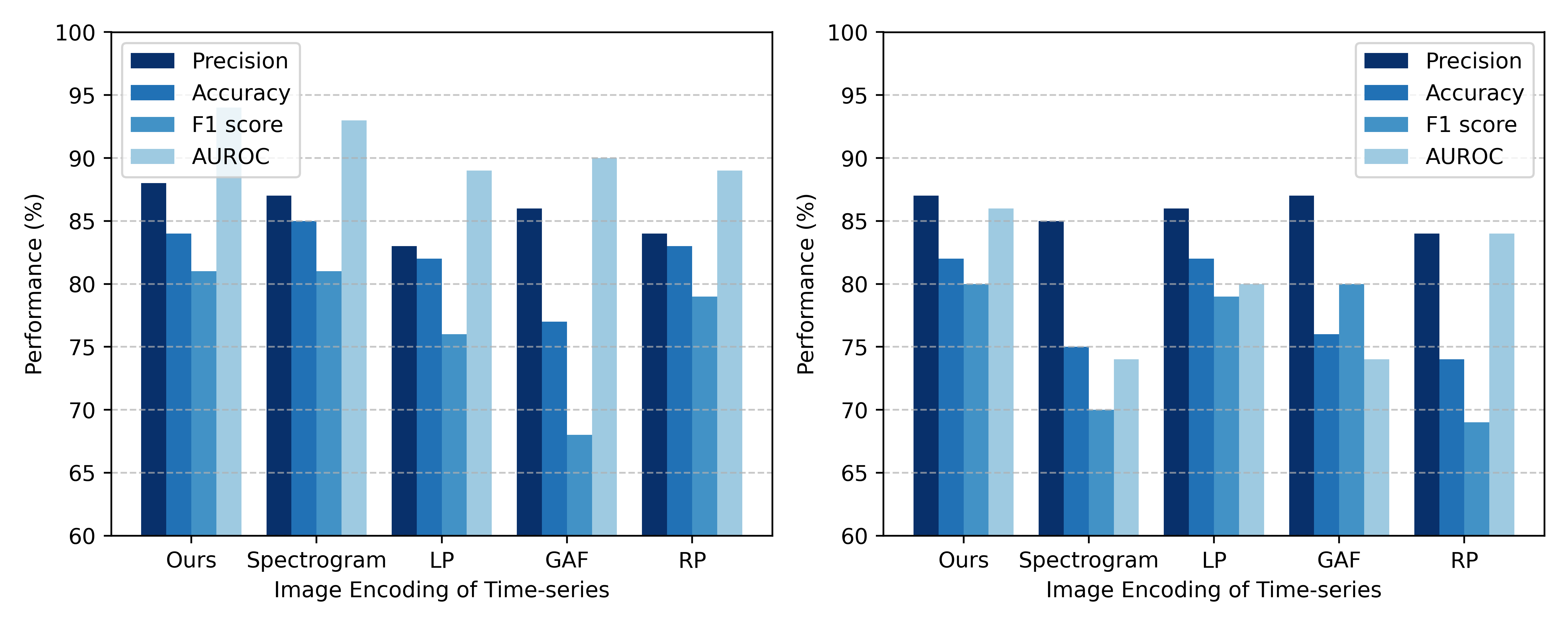}
    \caption{Comparison of time series imaging methods (Spectrogram, RP, GAF, and MTF) on PTB and APAVA datasets.}
    \label{fig:two_ts_images}
\vspace{-1em}
\end{figure}

\begin{figure}[!ht]
    \centering
    \includegraphics[width=0.90\linewidth]{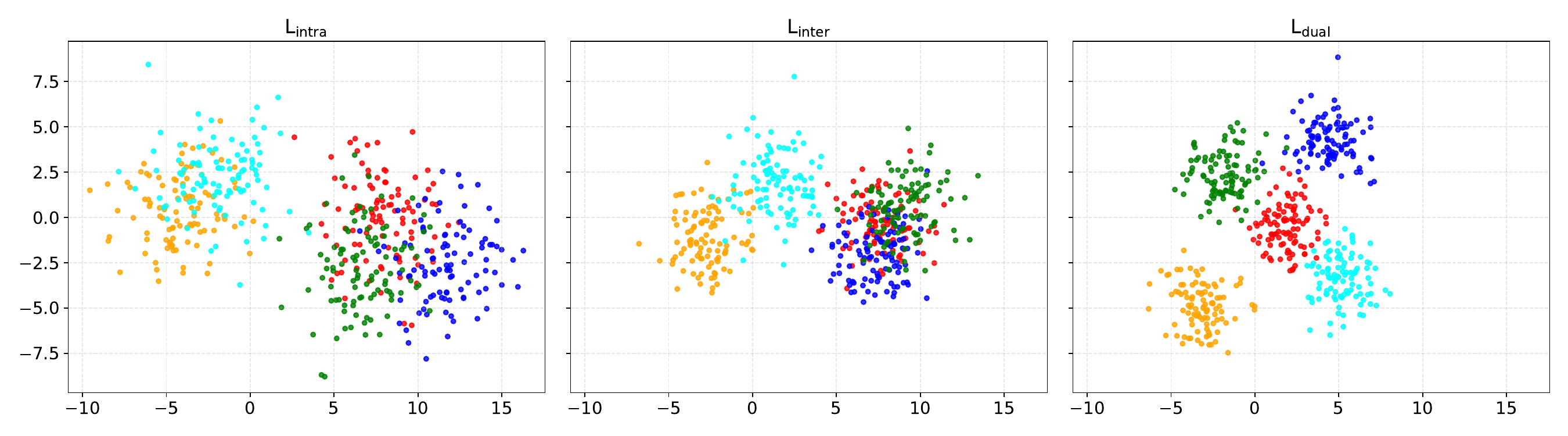}
    \caption{T-SNE visualization with different contrastive losses on PTB-XL.}
    \label{fig:three_tsne}
\vspace{-1em}
\end{figure}


\subsubsection{Domain Generalization Analysis}
Table~\ref{tab:seeg_domain} presents the subject-independent performance on sEEG dataset. Our method consistently improves accuracy across different architectures, rasing VanillaNet's from \(54.86\%\) to \(69.23\%\). Both cross- and intra-modal contrastive learning contribute, with visual knowledge providing notable gains.
As Table~\ref{table:seeg_domain_detail} shows, performance varies in pairwise cross-subject transfer: accuracy is high between similar subjects (\textit{e.g.}, P1 $\to$ P5), but drops for dissimilar pairs (\textit{e.g.}, P4 $\to$ P5), indicating substantial distributional shifts. 
These results show shat our method achieves strong generalization but faces limitations under extreme cross-subject variability.

\vspace{-1em}

\subsection{Ablation Studies}
\label{label:ablation_studies}
\vspace{-0.5em}

\subsubsection{Effectiveness of Dual Contrastive Learning}
We evaluate the dual contrastive learning on sEEG and three public datasets. As shown in Table~\ref{tab:seeg_subject_dependent}, adding the cross-modal loss $\mathcal{L}_{inter}$ improves accuracy from \(87.31\%\) to \(90.20\%\), outperforming the intra-modal loss $\mathcal{L}_{intra}$ \(+1.22\%\). Combining both losses achieves \(91.75\%\), confirming their complementary effect. Similar gains across public datasets (Fig.~\ref{fig:abl_dual_loss}) and larger improvements in less heterogeneous settings suggest that contrastive learning performs best under moderate subject variability.

\vspace{-1em}

\subsubsection{Model-agnostic Component Ablation}
As shown in Table~\ref{tab:seeg_subject_dependent} and Table~\ref{tab:seeg_domain}, our method improves performance across diverse vision models on sEEG under both protocols, demonstrating broad compatibility and robustness.

\vspace{-1em}

\subsubsection{Learnable \textit{vs.} Rule-Based Transformation}
Fig.~\ref{fig:two_ts_images} shows our learnable method outperforms Spectrogram, Recurrence Plot (RP), Gramian Angular Field (GAF), and Markov Transition Field (MTF) on PTB and APAVA. While rule-based fixed transformations that distort temporal structure, our adaptive approach preserves sequence integrity and aligns better with vision priors.

\subsection{T-SNE Visualization}
We apply t-SNE to visualize the learned embeddings on PTB-XL, illustrating the impact of different contrastive strategies. As shown in Fig.~\ref{fig:three_tsne}, the $\mathcal{L}_{intra}$ improves intra-class compactness, $\mathcal{L}_{inter}$ enhances cross-modal alignment, and the combined $\mathcal{L}_{dual}$ yields clearer and more discriminative embeddings, demonstrating that our method effectively captures robust and discriminative representations.

\section{Conclusion}
In this work, we propose {TS-P$^2$CL}, a plug-and-play framework for medical time series classification that enhances cross-subject generalization through vision-guided dual contrastive learning. By transforming 1D physiological signals into 2D pseudo-images,  our method bridges the modality gap and enables the transfer of universal visual priors. The dual contrastive objective harmonizes intra-modal consistency with cross-modal alignment to learn robust, domain-invariant representations, while keeping the  that are both temporally coherent and semantically enriched. The objective combines cross-modal alignment with pre-trained vision models and intra-modal consistency among augmented views, which strengthens the model’s ability to learn robust representations, while keeping the vision encoder frozen ensures plug-and-play efficiency. 
Experimental results on six physiological time series datasets (EEG, ECG, sEEG) demonstrate that our method achieves strong performance in subject- and subject-independent settings, with notable gains in cross-subject scenarios. 

%
%
%

\bibliographystyle{ieeetr}
\bibliography{references}
\end{document}